\PassOptionsToPackage{unicode}{hyperref}
\PassOptionsToPackage{hyphens}{url}
\PassOptionsToPackage{dvipsnames,svgnames,x11names}{xcolor}
\documentclass[
  10pt,
  letterpaper,
  twocolumn]{article}
\usepackage{xcolor}
\usepackage[margin=0.75in]{geometry}
\usepackage{amsmath,amssymb}

\newcommand{\Dkl}{D_{\mathrm{KL}}}
\usepackage{iftex}
\ifPDFTeX
  \usepackage[T1]{fontenc}
  \usepackage[utf8]{inputenc}
  \usepackage{textcomp} 
\else 
  \usepackage{unicode-math} 
  \defaultfontfeatures{Scale=MatchLowercase}
  \defaultfontfeatures[\rmfamily]{Ligatures=TeX,Scale=1}
\fi
\usepackage{times}
\ifPDFTeX\else
\fi
\IfFileExists{upquote.sty}{\usepackage{upquote}}{}
\IfFileExists{microtype.sty}{
  \usepackage[]{microtype}
  \UseMicrotypeSet[protrusion]{basicmath} 
}{}
\makeatletter
\@ifundefined{KOMAClassName}{
  \IfFileExists{parskip.sty}{%
    \usepackage{parskip}
  }{
    \setlength{\parindent}{0pt}
    \setlength{\parskip}{6pt plus 2pt minus 1pt}}
}{
  \KOMAoptions{parskip=half}}
\makeatother
\usepackage{color}
\usepackage{fancyvrb}
\usepackage{fvextra}

\DefineVerbatimEnvironment{Highlighting}{Verbatim}{commandchars=\\\{\},fontsize=\footnotesize,breaklines=true,breakanywhere=true}

\usepackage{longtable,booktabs,array}
\usepackage{calc} 
\usepackage{etoolbox}
\makeatletter
\patchcmd\longtable{\par}{\if@noskipsec\mbox{}\fi\par}{}{}
\makeatother
\IfFileExists{footnotehyper.sty}{\usepackage{footnotehyper}}{\usepackage{footnote}}
\makesavenoteenv{longtable}
\usepackage{graphicx}
\makeatletter
\newsavebox\pandoc@box
\newcommand*\pandocbounded[1]{
  \sbox\pandoc@box{#1}%
  \Gscale@div\@tempa{\textheight}{\dimexpr\ht\pandoc@box+\dp\pandoc@box\relax}%
  \Gscale@div\@tempb{\linewidth}{\wd\pandoc@box}%
  \ifdim\@tempb\p@<\@tempa\p@\let\@tempa\@tempb\fi
  \ifdim\@tempa\p@<\p@\scalebox{\@tempa}{\usebox\pandoc@box}%
  \else\usebox{\pandoc@box}%
  \fi%
}
\def\fps@figure{htbp}
\makeatother
\providecommand{\tightlist}{%
  \setlength{\itemsep}{0pt}\setlength{\parskip}{0pt}}
\usepackage[numbers,sort&compress]{natbib}
\usepackage{bookmark}
\IfFileExists{xurl.sty}{\usepackage{xurl}}{} 
\hypersetup{
  colorlinks=true,
  linkcolor={black},
  filecolor={black},
  citecolor={black},
  urlcolor={black},
  pdftitle={When Outcome Looks Right But Discipline Fails: Trace-Based Evaluation Under Hidden Competitor State},
  pdfauthor={Peiying Zhu, Sidi Chang},
  pdfcreator={LaTeX via pandoc}}

\title{\textbf{When Outcome Looks Right But Discipline Fails}\\
\large Trace-Based Evaluation Under Hidden Competitor State}
\author{
Peiying Zhu\\
Blossom AI\\
San Francisco, USA
\and
Sidi Chang\\
Blossom AI Labs\\
Tokyo, Japan
}
\date{}

\begin{document}

\maketitle

\begin{abstract}

Outcome-only evaluation can certify economically unsafe agents: a policy
can hit a business KPI while violating deployable behavioral discipline.
In hotel pricing with hidden competitor state, a learner can achieve
plausible revenue per available room while failing to preserve the rate
discipline of a rule-based revenue-management competitor. We introduce
discipline stability, a trace-based evaluation paradigm: define the
benchmark behavior, restrict observations to the deployment regime,
induce trace diagnostics from failure, separate mechanisms with
ablations, and test transfer and deployment. Across a two-hotel benchmark
and a compact hidden-budget bidding task, reward-only PPO variants miss
trace alignment; revealing hidden state reduces label uncertainty;
deterministic copy collapses uncertainty; and trace-prior or
corrected-history policies better preserve price or bid distributions.
Pure behavior cloning is nearly enough for symmetric imitation, while
Trace-Prior RL adds bounded adaptation under capacity asymmetry. The
contribution is an evaluation and benchmark paradigm, not a new optimizer
or a universal claim about MARL.

\end{abstract}
\vspace{0.1in}

\section{Introduction}\label{introduction}

Outcome-only benchmarks can certify the wrong behavior. This is
especially dangerous for strategic economic agents, where the outcome
metric is often a compressed business KPI and the intended behavior is a
discipline: yield management, budget pacing, queue balancing, routing
restraint, or compliance with a market protocol.

Hotel pricing makes the problem concrete. RevPAR is a natural outcome
metric, but it does not say how the agent earned revenue. A pricing
agent can sell too aggressively, undercut the benchmark, collapse to a
modal price, or distort the market it is supposed to learn from. Those
policies can look acceptable on revenue while failing the behavioral
logic that makes them deployable.

This paper asks a more demanding question, aligned with the
reward-hacking and reward-misspecification concerns studied by Skalse et
al.~(2022) and Pan et al. (2022):

\begin{quote}
Does the learned agent preserve the behavioral discipline of the
benchmark system after training, transfer, and deployment?
\end{quote}

We call this property discipline stability. The term is deliberately
empirical: we do not claim a universal theorem. The aim is a paradigm
other researchers can follow: define the intended discipline, expose
where scalar success is ambiguous, measure the trace of behavior, and
only then ask which learning method preserves that trace under the
deployable information regime.

Our testbed is a two-hotel pricing simulator. Hotel A is the learner.
Hotel B is a Fixed RM competitor using a deterministic rule based on
time, its own hidden remaining inventory, and market condition. Guests
choose Hotel A, Hotel B, or the outside option. When Hotel A undercuts,
it can redirect demand away from Hotel B, changing the data-generating
process it is trying to learn from.

The core difficulty is not that Hotel A has no context. In the
deployable sense, the agent is context-aware: it conditions on days
left, own remaining rooms, market condition, own booking pace, and, in
CA regimes, lagged market prices. But it is not fully
competitor-state-aware, because Hotel B's inventory and pricing rule
remain hidden. This partial context is exactly what makes the target
distributional.

\begin{figure*}[t]
\centering
\includegraphics[width=0.95\textwidth,keepaspectratio,alt={Hidden-state aliasing mechanism. The same visible Hotel A state can correspond to multiple hidden Hotel B inventories and therefore multiple valid market prices.}]{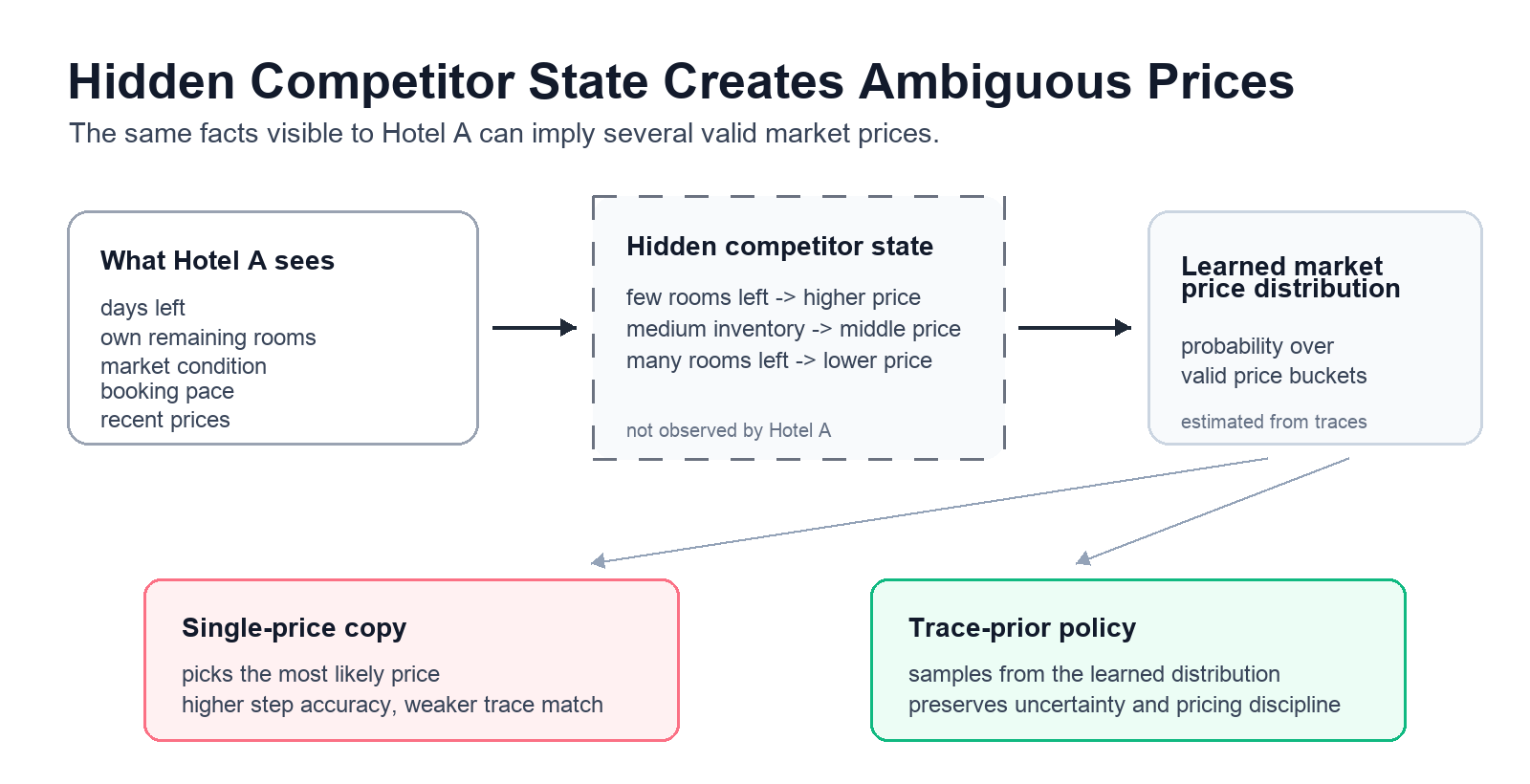}
\caption{Hidden-state aliasing mechanism. The same visible Hotel A state
can correspond to multiple hidden Hotel B inventories and therefore
multiple valid market prices.}
\end{figure*}

This is a trace-learning problem as much as an RL problem. Hotel A is
not given Hotel B's booking curve, pricing formula, or remaining
inventory. It only sees partial traces of a market process that depends
on variables it cannot observe. The learner therefore has to turn an
incomplete trace signal into deployable pricing behavior.

We build on the market-alignment result of Zhu and Chang (2026), which
showed that a Trace-Prior RL teacher can learn market-like pricing
against a Fixed RM competitor under hidden competitor state. Here we
broaden the question from market learning to discipline stability:
stronger reward-only baselines, student transfer, frozen learned-agent
deployment, state-sliced diagnostics, and second-domain mechanism
evidence.

The paper makes five claims that are supported by the current
experiments:

\begin{enumerate}
\def\labelenumi{\arabic{enumi}.}
\tightlist
\item
  Reward-only success is incomplete. PPO, recurrent PPO, and CTDE PPO
  are negative controls showing that scalar reward, memory, and hidden
  critic information do not recover Fixed-RM-like pricing traces in this
  benchmark.
\item
  Hidden competitor inventory is a major source of action-label
  uncertainty. Revealing oracle \(q_B\) sharply improves
  market-price prediction.
\item
  Trace learning is the repair signal. In the symmetric market, BC-only
  stochastic copy is nearly enough; in a capacity-asymmetric variant, a
  full-distribution Trace-Prior RL policy adds bounded objective
  adaptation while preserving most of the benchmark trace.
\item
  A student can internalize most of this discipline with less direct
  market dependence, and frozen deployment remains symmetric with and
  without lagged competitor-price context rather than collapsing into the
  early low-price shortcut.
\item
  The same paradigm reproduces in a second economic-agent POMDP:
  hidden-budget bidding, where outcome value can stay close while
  bid-distribution and pacing discipline fail.
\end{enumerate}

The paper does not claim that all MARL algorithms fail, that this
simulator is a real hotel market, or that frozen deployment proves
online co-learning safety. Those are future extensions. The current
contribution is the discipline-stability evaluation paradigm: a concrete
way to test whether strategic economic agents preserve benchmark
behavior rather than merely matching an outcome metric.

The claim boundary is therefore precise. We show that outcome-only
success is incomplete in the pricing benchmark, not that RevPAR should
be discarded. We show that hidden \(q_B\) explains much of the market-label
uncertainty, not that oracle \(q_B\) is available in deployment.
We show that a stochastic trace prior repairs this fixed-RM benchmark
better than reward-only negative controls and sampled-action trace
baselines, not that all modern MARL algorithms fail. We show that
corrected-history students can inherit
aggregate discipline and that frozen deployment does not immediately
collapse with or without lagged competitor-price context, not that
student trajectories exactly copy the teacher or that online co-learning
markets are solved. Hidden-budget
bidding supports the paradigm in a second economic-agent POMDP, but it
is still compact and synthetic; it is not a full cross-domain benchmark
suite.

\section{Environment and Information
Structure}\label{environment-and-information-structure}

Each episode is a finite-horizon selling problem with horizon
\texttt{H\ =\ 30} and capacity \texttt{Q\ =\ 100}. The action grid is
\(\mathcal{A}=\{100,120,140,160,180,200,220\}\). At each step, Hotel A
chooses a price \(p_{A,t}\) and Hotel B chooses
\(p_{B,t}=\mathrm{FixedRM}(t,q_{B,t},m_t)\),

where \(q_{B,t}\) is Hotel B's remaining inventory and \(m_t\)
is the market condition. The Fixed RM rule is a hand-tuned
revenue-management benchmark: it uses an OSG-style price rule plus a
booking-pace floor. It is deterministic, but Hotel A does not observe
its inputs.

Hotel B is not a learning agent in the main benchmark. Although guests
choose between Hotel A, Hotel B, and walking away, the training problem
is best viewed as a fixed-opponent single-agent POMDP from Hotel A's
perspective, not as a full co-learning MARL game. We keep the market
language because Hotel A's actions can change which hotel receives
demand and therefore alter the trace it observes from the fixed
competitor.

This is a partially observable decision problem in the standard sense of
POMDPs (Kaelbling et al., 1998). The full simulator state can be written
as \(s_t=(\tau_t,q_{A,t},q_{B,t},m_t,h_{A,t},h_{B,t},\text{price histories})\),

where \(\tau_t\) is time remaining, \(q_{A,t}\) and
\(q_{B,t}\) are remaining rooms, \(h_{A,t}\) and \(h_{B,t}\)
are booking-pace summaries, and price histories record recent realized
prices. Hotel A receives an observation \(o_{A,t}=O_r(s_t)\),

where the information regime \texttt{r} determines what is visible.

We use two main regimes:

{\def\LTcaptype{none} 
\begin{table*}[t]
\centering
\scriptsize
\begin{tabular}{@{}
  >{\raggedright\arraybackslash}p{(\linewidth - 4\tabcolsep) * \real{0.3333}}
  >{\raggedright\arraybackslash}p{(\linewidth - 4\tabcolsep) * \real{0.3333}}
  >{\raggedright\arraybackslash}p{(\linewidth - 4\tabcolsep) * \real{0.3333}}@{}}
\toprule\noalign{}
\begin{minipage}[b]{\linewidth}\raggedright
Regime
\end{minipage} & \begin{minipage}[b]{\linewidth}\raggedright
Deployable observation
\end{minipage} & \begin{minipage}[b]{\linewidth}\raggedright
What is hidden
\end{minipage} \\
\midrule\noalign{}
NC & own state plus own corrected price history & Hotel B price history
and Hotel B state \\
CA & own state plus lagged market/Hotel B prices & Hotel B inventory and
pricing rule \\
\end{tabular}
\end{table*}
}

The reward-only baselines optimize per-step RevPAR,
\(r_{A,t}=p_{A,t}y_{A,t}/Q\),

where \(y_{A,t}\) is Hotel A's sold room count at step \(t\).
This reward naturally encourages selling rooms, but by itself it does
not distinguish disciplined yield management from low-price occupancy
grabbing.

The POMDP trigger is that \(q_{B,t}\) is hidden from Hotel A.

Therefore the same observed state \(o_{A,t}\) can correspond to
several possible competitor inventories and several valid Fixed RM
prices. The market target is not one label. It is the posterior
predictive distribution
\(\pi_M(a\mid o_{A,t})\approx P(a_{B,t}=a\mid o_{A,t})\).

This distribution is estimated from traces. The purpose of the market
prior is not to reveal Hotel B's private state. It is to preserve the
uncertainty created by that missing state.

\section{Discipline Stability}\label{discipline-stability}

We define discipline stability as an empirical benchmark property. A
policy is discipline-stable relative to a benchmark if it preserves both
outcome and trace structure under the information regime in which it is
deployed.

More formally, let \(B\) be a benchmark policy, \(r\) a deployable
information regime, \(\mathcal{C}\) a set of state slices, and
\(\mathcal{Z}\) a set of
trace variables such as prices, sales, occupancy, ADR, and remaining
inventory. A learned policy \(\pi\) passes a discipline-stability check
at tolerances \((\epsilon_{\mathrm{out}},\epsilon_z)\) if its outcome gap
is small and its trace distributions are close to the benchmark both in
aggregate and on the chosen slices:
\[
\begin{aligned}
\Delta_{\mathrm{out}}(\pi,B;r) &\le \epsilon_{\mathrm{out}},\\
D(P_\pi(z\mid c,r),P_B(z\mid c,r)) &\le \epsilon_z,
\qquad \forall z\in \mathcal{Z},\; c\in \mathcal{C}.
\end{aligned}
\]
Here \(\Delta_{\mathrm{out}}\) is the absolute difference in expected
scalar outcome under regime \(r\); in pricing it is
\(|\mathbb{E}_{r,\pi}[\mathrm{RevPAR}]
-\mathbb{E}_{r,B}[\mathrm{RevPAR}]|\). For action-distribution diagnostics
we instantiate \(D\) with
\(D_1(P,Q)=\sum_x |P(x)-Q(x)|\) and Jensen-Shannon divergence
\[
\begin{aligned}
D_{\mathrm{JS}}(P,Q)=
\frac{1}{2}\Dkl(P\|M)+\frac{1}{2}\Dkl(Q\|M),\\
M=(P+Q)/2 .
\end{aligned}
\]

The thresholds are domain choices, not universal constants. The
contribution is the evaluation form: a strategic agent should not be
certified only because its scalar KPI is close.

The proposed paradigm has five steps.

\begin{enumerate}
\def\labelenumi{\arabic{enumi}.}
\tightlist
\item
  Define the benchmark discipline. State what the reference system is
  meant to preserve beyond outcome: rate discipline in pricing, pacing
  in bidding, balance in routing, or compliance in a protocol.
\item
  Define the information regime. Write down what the deployed agent
  observes and which variables remain hidden. This prevents evaluation
  from silently grading the agent against information it cannot have.
\item
  Induce trace diagnostics from the failure. Start with the scalar KPI,
  then decompose how it was achieved: business components, action
  distributions, state slices, seed-level uncertainty, and deployment
  interactions.
\item
  Diagnose before repairing. Use an ablation ladder to separate weak
  optimization, missing memory, hidden-state ambiguity, deterministic
  collapse, and distributional repair.
\item
  Test persistence. After the benchmark signal is reduced or removed,
  test whether the learned discipline transfers to a student and remains
  stable in interaction.
\end{enumerate}

This is the main methodological contribution. Trace-Prior RL is one
repair that passes the protocol in this benchmark; it is not the
paradigm itself.

The findings translate into a reusable protocol. If the scalar KPI
passes but the trace fails, the benchmark is not solved; the evaluator
must report how the KPI was produced. If a hidden variable makes one
visible state compatible with several benchmark actions, the evaluator
should treat the benchmark target as a distribution, not as a single
correct action. If deterministic copying has high step accuracy but poor
trace alignment, the evaluator should test stochastic repairs that
preserve ambiguity. If the repair only works while the benchmark signal
is present, the evaluator should run a transfer or deployment test after
that signal is weakened.

The minimal reporting checklist is therefore concrete:

\begin{itemize}
\tightlist
\item
  Benchmark discipline: what behavior should survive beyond the scalar
  KPI?
\item
  Deployed observation: what does the agent actually see?
\item
  Hidden state: which benchmark-relevant variables are missing?
\item
  Scalar outcome: does the agent pass the normal KPI?
\item
  Trace diagnostics: does it match action distributions, state slices,
  and business components?
\item
  Mechanism ablations: does the failure come from optimization, hidden
  state, deterministic collapse, or repair choice?
\item
  Persistence tests: does the discipline survive after the benchmark
  signal is reduced or removed?
\end{itemize}

The diagnostics are induced from the failure. RevPAR answers whether the
agent earned revenue, but not whether it earned revenue through
deployable yield management. Once the failure is visible, the missing
evidence is specific: occupancy tells us whether the agent is buying
revenue through excess selling; ADR tells us whether it preserves rate
discipline; price-bucket distributions tell us whether the policy uses
the market's action structure; state slices tell us whether aggregate
alignment hides local failure.

Let \(B\) be the benchmark policy or trace reference, \(r\) an
information regime, \(z\) a trace variable, and \(c\) a state slice. In
this paper, \(z\) includes price bucket, occupancy, ADR, remaining
inventory, and sales. We measure outcome gap
\(\Delta_{\mathrm{out}}(\pi,B;r)\), trace distance
\(D(P_\pi(z\mid c,r),P_B(z\mid c,r))\), transfer distance
\(D(P_{\mathrm{student}}(z\mid c),P_{\mathrm{teacher}}(z\mid c))\), and
interaction gap \(\Delta_{\mathrm{pair}}(\pi_i,\pi_j)\).

The exact thresholds are not universal. The point is the evaluation
protocol: RevPAR alone is necessary but insufficient. We report:

{\def\LTcaptype{none} 
\begin{table*}[t]
\centering
\scriptsize
\begin{tabular}{@{}
  >{\raggedright\arraybackslash}p{(\linewidth - 4\tabcolsep) * \real{0.3333}}
  >{\raggedright\arraybackslash}p{(\linewidth - 4\tabcolsep) * \real{0.3333}}
  >{\raggedright\arraybackslash}p{(\linewidth - 4\tabcolsep) * \real{0.3333}}@{}}
\toprule\noalign{}
\begin{minipage}[b]{\linewidth}\raggedright
Layer
\end{minipage} & \begin{minipage}[b]{\linewidth}\raggedright
Question
\end{minipage} & \begin{minipage}[b]{\linewidth}\raggedright
Metrics
\end{minipage} \\
\midrule\noalign{}
Outcome & Did the agent earn revenue? & RevPAR \\
Business decomposition & How did it earn revenue? & occupancy, ADR \\
Trace behavior & Did it price like the benchmark? & price buckets, L1,
JS \\
Uncertainty & Is the result stable across seeds? & seed-level 95\% CI \\
Conditional behavior & Is aggregate match hiding local failure? &
state-sliced L1/JS \\
Deployment & Does discipline survive interaction? & NC-vs-NC, CA-vs-CA,
NC-vs-CA \\
\end{tabular}
\end{table*}
}

\section{Negative Controls and Information-Symmetric
Baselines}\label{negative-controls-and-information-symmetric-baselines}

A natural alternative explanation is that earlier DQN results failed
only because the optimizer was weak. We therefore compare against
PPO (Schulman et al., 2017), recurrent PPO, and CTDE PPO, which is
motivated by the centralized training/decentralized execution line of
multi-agent actor-critic work (Lowe et al., 2017). All three optimize
only \(r_{A,t}=p_{A,t}y_{A,t}/Q\).

There is no teacher, no market prior, and no KL regularizer. Recurrent
PPO tests whether memory solves partial observability. CTDE PPO gives
the critic hidden competitor information during training, while the
deployable actor remains decentralized.

These reward-only runs are negative controls, not information-symmetric
competitors to Trace-Prior RL. They ask whether scalar reward, memory,
or hidden critic information is enough. A fair trace-access comparison
must also give the baseline access to Fixed-RM trace labels.

The ablation logic is deliberately sequential. Reward-only PPO tests
whether DQN was simply too weak. Recurrent PPO tests whether generic
memory is enough. CTDE PPO tests whether hidden training-time
information in the critic is enough when the deployed actor still
optimizes only reward. The oracle-\(q_B\) predictor tests whether
hidden state is actually the trigger. The argmax predictor tests whether
deterministic copying is enough. Trace-Prior RL tests whether preserving
the full market mixture repairs the trace. The corrected-history student
tests whether that discipline can transfer after direct market-price
dependence is reduced.
For compact tables, R-PPO denotes recurrent PPO and Student denotes the
corrected-history student.

{\def\LTcaptype{none} 
\begin{table*}[t]
\centering
\scriptsize
\begin{tabular}{@{}
  >{\raggedright\arraybackslash}p{(\linewidth - 18\tabcolsep) * \real{0.1450}}
  >{\raggedleft\arraybackslash}p{(\linewidth - 18\tabcolsep) * \real{0.0950}}
  >{\raggedleft\arraybackslash}p{(\linewidth - 18\tabcolsep) * \real{0.0950}}
  >{\raggedleft\arraybackslash}p{(\linewidth - 18\tabcolsep) * \real{0.0950}}
  >{\raggedleft\arraybackslash}p{(\linewidth - 18\tabcolsep) * \real{0.0950}}
  >{\raggedleft\arraybackslash}p{(\linewidth - 18\tabcolsep) * \real{0.0950}}
  >{\raggedleft\arraybackslash}p{(\linewidth - 18\tabcolsep) * \real{0.0950}}
  >{\raggedleft\arraybackslash}p{(\linewidth - 18\tabcolsep) * \real{0.0950}}
  >{\raggedleft\arraybackslash}p{(\linewidth - 18\tabcolsep) * \real{0.0950}}
  >{\raggedleft\arraybackslash}p{(\linewidth - 18\tabcolsep) * \real{0.0950}}@{}}
\toprule\noalign{}
\begin{minipage}[b]{\linewidth}\raggedright
Method
\end{minipage} & \begin{minipage}[b]{\linewidth}\raggedleft
Seeds
\end{minipage} & \begin{minipage}[b]{\linewidth}\raggedleft
RevPAR A
\end{minipage} & \begin{minipage}[b]{\linewidth}\raggedleft
RevPAR B
\end{minipage} & \begin{minipage}[b]{\linewidth}\raggedleft
Occ A
\end{minipage} & \begin{minipage}[b]{\linewidth}\raggedleft
Occ B
\end{minipage} & \begin{minipage}[b]{\linewidth}\raggedleft
ADR A
\end{minipage} & \begin{minipage}[b]{\linewidth}\raggedleft
ADR B
\end{minipage} & \begin{minipage}[b]{\linewidth}\raggedleft
L1
\end{minipage} & \begin{minipage}[b]{\linewidth}\raggedleft
JS
\end{minipage} \\
\midrule\noalign{}
PPO & 10 & 93.554 & 106.925 & 0.7342 & 0.7964 & 127.68 &
134.21 & 0.4635 & 0.0559 \\
R-PPO & 10 & 95.026 & 101.202 & 0.7787 & 0.7677 &
122.30 & 131.76 & 0.5400 & 0.0657 \\
CTDE PPO & 10 & 94.482 & 110.182 & 0.7377 & 0.8000 & 128.26
& 137.71 & 0.6582 & 0.0959 \\
Trace-Prior & 10 & 108.063 & 107.931 & 0.7705 & 0.7677 & 140.26
& 140.59 & 0.0165 & 0.0000 \\
Student & 10 & 107.588 & 108.683 & 0.7670 & 0.7726 &
140.29 & 140.68 & 0.0198 & 0.0001 \\
\end{tabular}
\end{table*}
}

The reward-only baselines fail in two ways. First, they do not match the
business profile of Fixed RM: ADR and occupancy move in the wrong
combinations. Second, their price distributions are far from the
benchmark. This is exactly the discipline-stability failure: the scalar
objective does not recover the behavioral structure of the system the
agent enters.

{\def\LTcaptype{none} 
\begin{table*}[t]
\centering
\scriptsize
\begin{tabular}{@{}
  >{\raggedright\arraybackslash}p{(\linewidth - 14\tabcolsep) * \real{0.1450}}
  >{\raggedleft\arraybackslash}p{(\linewidth - 14\tabcolsep) * \real{0.1220}}
  >{\raggedleft\arraybackslash}p{(\linewidth - 14\tabcolsep) * \real{0.1220}}
  >{\raggedleft\arraybackslash}p{(\linewidth - 14\tabcolsep) * \real{0.1220}}
  >{\raggedleft\arraybackslash}p{(\linewidth - 14\tabcolsep) * \real{0.1220}}
  >{\raggedleft\arraybackslash}p{(\linewidth - 14\tabcolsep) * \real{0.1220}}
  >{\raggedleft\arraybackslash}p{(\linewidth - 14\tabcolsep) * \real{0.1220}}
  >{\raggedleft\arraybackslash}p{(\linewidth - 14\tabcolsep) * \real{0.1220}}@{}}
\toprule\noalign{}
\begin{minipage}[b]{\linewidth}\raggedright
Method
\end{minipage} & \begin{minipage}[b]{\linewidth}\raggedleft
100
\end{minipage} & \begin{minipage}[b]{\linewidth}\raggedleft
120
\end{minipage} & \begin{minipage}[b]{\linewidth}\raggedleft
140
\end{minipage} & \begin{minipage}[b]{\linewidth}\raggedleft
160
\end{minipage} & \begin{minipage}[b]{\linewidth}\raggedleft
180
\end{minipage} & \begin{minipage}[b]{\linewidth}\raggedleft
200
\end{minipage} & \begin{minipage}[b]{\linewidth}\raggedleft
220
\end{minipage} \\
\midrule\noalign{}
PPO & 13.19\% & 25.25\% & 29.82\% & 10.59\% & 11.35\% &
5.54\% & 4.25\% \\
R-PPO & 21.18\% & 30.09\% & 21.92\% & 11.06\% &
7.25\% & 5.43\% & 3.06\% \\
CTDE PPO & 22.57\% & 19.38\% & 16.05\% & 15.06\% & 11.04\% &
8.42\% & 7.48\% \\
Trace-Prior & 4.76\% & 43.19\% & 22.87\% & 14.89\% & 10.94\% &
3.34\% & 0.00\% \\
Student & 4.94\% & 42.48\% & 22.30\% & 15.36\% &
11.07\% & 3.84\% & 0.00\% \\
Hotel B & 4.91\% & 42.68\% & 22.56\% & 15.16\% & 11.29\% & 3.39\%
& 0.01\% \\
\end{tabular}
\end{table*}
}

We therefore added information-symmetric trace baselines. BC-only
stochastic copying trains a supervised market-price model from Fixed-RM
traces and samples from it, with no RL objective. BC warm-start
initializes the PPO actor from the same labels, then fine-tunes with
RevPAR reward. PPO + BC auxiliary keeps the RevPAR reward but adds a
sampled-action trace loss,
\(L=L_{\mathrm{PPO}}+\alpha\,\mathrm{CE}(\pi_\theta(\cdot\mid o_t),a_{B,t})\).

This gives PPO access to the same kind of trace labels, but only as a
sampled-action cross-entropy term. It does not constrain the full action
distribution to the learned market prior.

{\def\LTcaptype{none}
\begin{table*}[t]
\centering
\scriptsize
\resizebox{\textwidth}{!}{%
\begin{tabular}{@{}llrrrrrrrr@{}}
\toprule\noalign{}
Method & Trace access & RevPAR A & RevPAR B & Occ A & Occ B & ADR A & ADR B & L1 & JS \\
\midrule\noalign{}
reward-only PPO & none & 93.554 & 106.925 & 0.7342 & 0.7964 & 127.68 & 134.21 & 0.4635 & 0.0559 \\
BC-only stochastic copy & trace labels, no RL & 107.930 & 107.734 & 0.7688 & 0.7655 & 140.40 & 140.74 & 0.0147 & 0.00004 \\
BC warm-start + PPO & trace pretraining & 96.142 & 109.222 & 0.7053 & 0.8072 & 136.74 & 135.20 & 0.3864 & 0.0265 \\
PPO + BC aux, alpha=0.1 & sampled-action loss & 98.033 & 100.768 & 0.7721 & 0.7715 & 126.99 & 130.61 & 0.0522 & 0.0059 \\
PPO + BC aux, alpha=1.0 & stronger sampled-action loss & 98.831 & 98.488 & 0.7864 & 0.7603 & 125.68 & 129.54 & 0.0254 & 0.0015 \\
BC warm-start + PPO + BC aux & pretraining + sampled-action loss & 98.330 & 101.136 & 0.7699 & 0.7716 & 127.75 & 131.06 & 0.0547 & 0.0034 \\
Trace-Prior teacher & full-distribution prior + RL & 108.063 & 107.931 & 0.7705 & 0.7677 & 140.26 & 140.59 & 0.0165 & 0.0000 \\
\bottomrule\noalign{}
\end{tabular}
}
\end{table*}
}

The BC-only row deserves to be read directly, not explained away. In
this default, roughly symmetric two-hotel setting, sampling from a
well-fit market prior is already almost enough: BC-only and the
Trace-Prior teacher are tied within seed noise on the main business and
distributional metrics. This is exactly why the algorithmic claim should
stay modest. The evidence says that the trace prior carries most of the
market discipline; the RL term is useful only when Hotel A must preserve
that discipline while optimizing an objective that is not exactly
``copy the benchmark.''

We ran a small asymmetric stress test to check this boundary. Hotel B and
the demand process stay fixed, but Hotel A's objective becomes
\((p_{A,t}-100)y_{A,t}/Q\), making low-price sales less valuable for Hotel
A. BC-only still tracks Hotel B very closely. Trace-Prior RL with
\(\beta=30\) increases Hotel A's objective RevPAR from \(30.964\) to
\(31.222\), keeps A-B L1 nearly unchanged but slightly worse (\(0.0098\) to
\(0.0102\)), and increases occupancy by \(+0.0051\) in a paired seed
comparison. The small trace-shape cost is the expected price of allowing
Hotel A to re-weight its own objective instead of copying the benchmark
exactly. The objective gain is directional rather than statistically
conclusive with five seeds: paired 95\% CI \([-0.320,+0.836]\), which
crosses zero. We therefore treat this as a boundary result, not as a
decisive win over behavior cloning, and return to this calibration in
Section~\ref{limitations}.

\subsection{Capacity-Asymmetric Stress
Test}\label{capacity-asymmetric-stress-test}

We then ran the cleaner capacity-asymmetric test that the default
benchmark was missing. Hotel B remains the same Fixed RM hotel with
capacity \(Q_B=100\), but Hotel A has larger capacity, \(Q_A=120\), and
its RevPAR is computed per available Hotel A room. In this setting,
exact copying is too conservative for the larger hotel.

\begin{table}[t]
\centering
\scriptsize
\resizebox{\columnwidth}{!}{%
\begin{tabular}{@{}lrrrrr@{}}
\toprule\noalign{}
Method & \(Q_A\) & RevPAR A & Occ A & ADR A & A-B L1 \\
\midrule\noalign{}
BC-only stochastic copy & 120 & 91.281 & 0.6475 & 140.97 & 0.0110 \\
Trace-Prior RL, \(\beta=30\) & 120 & 92.045 & 0.6527 & 141.03 & 0.0155 \\
\bottomrule\noalign{}
\end{tabular}
}
\caption{Capacity-asymmetric stress test. Hotel B keeps capacity \(Q_B=100\);
Hotel A has larger capacity \(Q_A=120\).}
\end{table}

The paired RevPAR improvement is \(+0.764\) with 95\% CI
\([+0.125,+1.403]\), so this CI does not cross zero. The paired occupancy
gain is also positive: \(+0.0051\) with 95\% CI \([+0.0014,+0.0088]\). In
business terms, the larger Hotel A fills more rooms at nearly the same
rates rather than charging more per room, which is the natural way to use
additional capacity while staying close to the benchmark. The trace-shape
cost is small: paired A-B L1 increases by only \(+0.0013\), with 95\% CI
\([-0.0006,+0.0032]\). Smaller-capacity variants (\(Q_A=80\) and \(Q_A=60\))
did not produce clean separation because BC-only already fills most of
Hotel A's inventory. This gives the most useful bounded algorithmic claim:
BC-only is enough for pure imitation, while Trace-Prior RL adds value when
the agent must preserve a benchmark discipline and adapt it to a different
deployable capacity.

These results narrow the interpretation. The claim is not that
supervised distribution matching beats reward-only RL on a distributional
metric. The supported claim is that discipline stability requires an
explicit trace-learning signal, and that the form of the trace signal
matters. A warm-start can be forgotten during reward fine-tuning. A
sampled-action auxiliary can make paired A-B price buckets look closer,
but with
\texttt{alpha=1.0} it does so in a lower-ADR paired market: Hotel A
sells more at lower rates, and the paired Hotel B trace also shifts
downward. The full-distribution prior preserves the offline benchmark
mixture while still allowing an RL policy to optimize Hotel A's reward.

The supported claim is narrow but important: in this benchmark, scalar
reward alone does not produce discipline-stable pricing under hidden
competitor state. Trace access helps, but sampled-action trace losses
and warm-starts are weaker than a full-distribution prior.

We do not claim that every modern MARL method would fail.

\section{Why the Target Is
Distributional}\label{why-the-target-is-distributional}

The failure is caused by hidden-state aliasing. The same Hotel A-visible
state can correspond to several Hotel B inventories, and therefore to
several valid Fixed RM prices:
\(o_{A,t}\rightarrow\{q_{B,t}\}\rightarrow\{a_{B,t}\}\).

This is the point at which the trace-learning view matters. The observed
Hotel B price is not a clean label for ``the right action in this
state.'' It is a sample from a market process whose latent state is
partly missing. The correct object to learn is therefore not only a class
label, but the observable posterior distribution over market actions.

We test this directly with two diagnostics.

First, replaying the final Trace-Prior policy over 5 seeds and 2,000
evaluation episodes per seed, we group visited steps into coarse Hotel
A-visible cells. If the observable state fully determined Hotel B's
action, these cells would be nearly single-action.

{\def\LTcaptype{none} 
\begin{table*}[t]
\centering
\scriptsize
\begin{tabular}{@{}lr@{}}
\toprule\noalign{}
Diagnostic & Value \\
\midrule\noalign{}
Total visited steps & 300,000 \\
Eligible cells, at least 30 samples & 1,951 \\
Eligible step share & 60.78\% \\
Cells with \textgreater=2 observed Hotel B actions & 95.08\% \\
Cells with \textgreater=2 actions each at \textgreater=5\% share &
85.29\% \\
Eligible steps in substantive ambiguous cells & 76.57\% \\
Weighted normalized within-cell entropy & 0.2800 \\
\end{tabular}
\end{table*}
}

Second, we train two supervised predictors for Hotel B's current price
bucket: one using only Hotel A-visible CA-lag3 features, and
one using the same features plus oracle \(q_B/Q\). No RL training
is performed in this diagnostic.

{\def\LTcaptype{none} 
\begin{table*}[t]
\centering
\scriptsize
\begin{tabular}{@{}
  >{\raggedright\arraybackslash}p{(\linewidth - 10\tabcolsep) * \real{0.1304}}
  >{\raggedleft\arraybackslash}p{(\linewidth - 10\tabcolsep) * \real{0.1739}}
  >{\raggedleft\arraybackslash}p{(\linewidth - 10\tabcolsep) * \real{0.1739}}
  >{\raggedleft\arraybackslash}p{(\linewidth - 10\tabcolsep) * \real{0.1739}}
  >{\raggedleft\arraybackslash}p{(\linewidth - 10\tabcolsep) * \real{0.1739}}
  >{\raggedleft\arraybackslash}p{(\linewidth - 10\tabcolsep) * \real{0.1739}}@{}}
\toprule\noalign{}
\begin{minipage}[b]{\linewidth}\raggedright
Predictor
\end{minipage} & \begin{minipage}[b]{\linewidth}\raggedleft
NLL
\end{minipage} & \begin{minipage}[b]{\linewidth}\raggedleft
Accuracy
\end{minipage} & \begin{minipage}[b]{\linewidth}\raggedleft
Brier
\end{minipage} & \begin{minipage}[b]{\linewidth}\raggedleft
True-price prob.
\end{minipage} & \begin{minipage}[b]{\linewidth}\raggedleft
Norm. entropy
\end{minipage} \\
\midrule\noalign{}
observable CA-lag3 & 0.5359 & 76.91\% & 0.3160 & 0.6865 &
0.2748 \\
oracle CA-lag3 + \(q_B\) & 0.1557 & 95.47\% & 0.0827 & 0.8867 &
0.1210 \\
\end{tabular}
\end{table*}
}

Revealing hidden competitor inventory makes the market label much
sharper. This does not make oracle \(q_B\) deployable. It supports
the causal story: the uncertainty is not merely noise or weak
optimization; it is induced by missing competitor state.

This also explains why deterministic copying is insufficient. A
deterministic argmax predictor can have higher step-level accuracy while
distorting the aggregate price trace, because it collapses the posterior
predictive mixture onto the modal bucket. Under hidden state, exact
action accuracy and trace alignment can move in opposite directions. The
repair must preserve uncertainty that the deployed agent cannot resolve.

\section{Trace-Prior Teacher}\label{trace-prior-teacher}

The teacher solves the market-learning stage. It does not receive the
hidden RM rule. Instead, it turns weak market traces into an explicit
intermediate structure: a distribution over the benchmark's likely
prices. The teacher sees deployable CA information, namely Hotel A's own
state and the last three Hotel B prices. It does not see Hotel B's
current inventory or rule. Teacher observation is
\[
o_{T,t}=
(\tau_t/H,\;q_{A,t}/Q,\;m_t,\;h_{A,t},\;a_{B,t-1},\;a_{B,t-2},\;a_{B,t-3}) .
\]

First, a supervised market prior estimates
\(\pi_M(a\mid o_{T,t})\approx P(a_{B,t}=a\mid o_{T,t})\).

Then the teacher policy is trained with a full-distribution KL regularizer
(Kullback and Leibler, 1951):
\[
\begin{aligned}
J(\theta)
&=\mathbb{E}_{\xi\sim\pi_\theta}\Bigg[
\sum_{t=0}^{H-1}\Big(
r^{\mathrm{out}}_{A,t}\\
&\quad-\beta\,\Dkl\!\big(
\pi_\theta(\cdot\mid o_{T,t})\,\big\|\,
\hat{\pi}_M(\cdot\mid o_{T,t})\big)
\Big)\Bigg].
\end{aligned}
\]
Here \(\xi\) denotes the trajectory induced by \(\pi_\theta\) and
\(\gamma=1\) over the finite horizon.

In plain language, the teacher is still an RL pricing agent: it earns
reward for Hotel A revenue. But every update also pays a cost if the
whole action distribution drifts away from the learned market trace. The
first term rewards selling at revenue-generating prices. The KL term
prevents the policy from inventing a pricing distribution that no longer
resembles the market discipline it is supposed to learn.

The coefficient \(\beta\) controls the interpolation. When
\(\beta=0\), the agent is a reward-only pricing policy. As
\(\beta\) grows, the policy is pulled closer to \(\pi_M\), the
empirical estimate of \(P(a_{B,t}\mid o_{T,t})\). The paper
uses this term as a discipline prior, not as a claim that \(\beta\)
is universally optimal.
We use the forward direction \(\Dkl(\pi_\theta\|\hat{\pi}_M)\), the
standard prior-regularized direction for penalizing action mass unsupported by
the trace prior; reversing the KL would impose a different, more
mode-covering constraint.

The final teacher uses \texttt{beta\ =\ 30} and no entropy bonus. This
matters because the stochasticity comes from the learned market prior,
not from an added entropy bonus.

Across 10 seeds, the Trace-Prior teacher closely matches the paired
Hotel B business metrics and price distribution: RevPAR 108.063 versus
107.931, occupancy 0.7705 versus 0.7677, ADR 140.26 versus 140.59,
price L1 0.0165, and JS approximately 0.

This is the strongest current repair for the teacher-market stage.

\section{Corrected-History
Student}\label{corrected-history-student}

The teacher still uses lagged Hotel B prices. The student asks a harder
deployment question: can the learned discipline survive after direct
market dependence is reduced? The student is designed for a more
independent regime. It uses Hotel A's own state and its own recent
corrected price history:
\[
\begin{aligned}
o_{S,t}=(&\tau_t/H,\;q_{A,t}/Q_A,\;m_t,\;h_{A,t},\\
&a_{A,t-1},\;a_{A,t-2},\;a_{A,t-3}) .
\end{aligned}
\]
When \(t-k<0\) for \(k\in\{1,2,3\}\), the missing lag is padded with a fixed
start-of-episode token encoded outside the price grid.

The teacher labels student-visible states with hard correction labels:
\[
\begin{aligned}
\mathcal{D} &= \{(o_{S,t},a_{T,t})\},\\
L_{\mathrm{student}}(\phi)
&= -\mathbb{E}_{(o_{S,t},a_{T,t})}
\log \pi_\phi(a_{T,t}\mid o_{S,t}) .
\end{aligned}
\]

This is corrected-history DAgger, adapting the dataset-aggregation idea
of Ross et al.~(2011). The student first learns from teacher
demonstrations. Then the student rolls forward using its own histories,
and the teacher labels the states the student actually creates. The
teaching signal is therefore not ``copy this one pristine trajectory.''
It is ``when your own history puts you in this state, choose the
disciplined price the teacher would choose.'' In one sentence: the
teacher corrects the student's practice rounds so the student learns how
to act under the histories it will actually create.

At deployment, the teacher is removed. The student samples from its
learned policy using only student-visible information.

The practical cost is teacher queries. During corrected-history
training, the teacher must label states induced by student rollouts. In
our simulator this is cheap because the teacher is a frozen policy. In
an industry deployment, the same idea would need a finite correction
window or an active-learning budget: query the teacher most often where
the student is uncertain or where the state is business-critical, then
remove the teacher after validation.

The selected student passes the aggregate good-student test:

{\def\LTcaptype{none} 
\begin{table*}[t]
\centering
\scriptsize
\begin{tabular}{@{}
  >{\raggedright\arraybackslash}p{(\linewidth - 8\tabcolsep) * \real{0.1667}}
  >{\raggedleft\arraybackslash}p{(\linewidth - 8\tabcolsep) * \real{0.2222}}
  >{\raggedleft\arraybackslash}p{(\linewidth - 8\tabcolsep) * \real{0.2222}}
  >{\raggedright\arraybackslash}p{(\linewidth - 8\tabcolsep) * \real{0.1667}}
  >{\raggedleft\arraybackslash}p{(\linewidth - 8\tabcolsep) * \real{0.2222}}@{}}
\toprule\noalign{}
\begin{minipage}[b]{\linewidth}\raggedright
Metric
\end{minipage} & \begin{minipage}[b]{\linewidth}\raggedleft
Student mean and 95\% CI
\end{minipage} & \begin{minipage}[b]{\linewidth}\raggedleft
Teacher mean
\end{minipage} & \begin{minipage}[b]{\linewidth}\raggedright
Teacher inside student CI?
\end{minipage} & \begin{minipage}[b]{\linewidth}\raggedleft
Gap
\end{minipage} \\
\midrule\noalign{}
RevPAR & 107.960 {[}107.026, 108.894{]} & 108.060 & yes & -0.100 \\
Occupancy & 0.7728 {[}0.7604, 0.7852{]} & 0.7703 & yes & +0.0025 \\
ADR & 139.71 {[}138.46, 140.95{]} & 140.29 & yes & -0.58 \\
\end{tabular}
\end{table*}
}

Its aggregate price distribution also matches both the teacher and Hotel
B:

{\def\LTcaptype{none} 
\begin{table*}[t]
\centering
\scriptsize
\begin{tabular}{@{}lrrrrrrr@{}}
\toprule\noalign{}
Policy & 100 & 120 & 140 & 160 & 180 & 200 & 220 \\
\midrule\noalign{}
Student & 4.66\% & 43.45\% & 22.56\% & 15.27\% & 10.63\% & 3.42\% &
0.00\% \\
Teacher & 4.64\% & 43.46\% & 22.54\% & 15.07\% & 10.69\% & 3.60\% &
0.00\% \\
Hotel B & 4.84\% & 42.31\% & 23.03\% & 15.37\% & 11.15\% & 3.29\% &
0.01\% \\
\end{tabular}
\end{table*}
}

The interpretation is strong habit transfer, not exact path-by-path
copying. Seed-level student-teacher price L1 is not zero, and this
should remain visible in the paper. The student has
internalized the teacher's discipline at the aggregate deployment level.

\section{Frozen Deployment Test}\label{frozen-deployment-test}

The next question is whether the learned policies interact sensibly
after the Fixed RM benchmark is removed. This is not online co-learning.
Policies are frozen, and no training is performed.

We test:

{\def\LTcaptype{none} 
\begin{table*}[t]
\centering
\scriptsize
\begin{tabular}{@{}ll@{}}
\toprule\noalign{}
Label & Meaning \\
\midrule\noalign{}
NC & corrected-history student; own state plus last three own prices \\
CA & Trace-Prior teacher; own state plus last three rival prices \\
\end{tabular}
\end{table*}
}

The deployment matrix asks whether the learned policies remain symmetric
in same-policy matchups and whether lagged competitor-price context
creates a directional mixed-matchup difference. It is a stability
diagnostic, not a proof of strategic equilibrium.

{\def\LTcaptype{none} 
\begin{table*}[t]
\centering
\scriptsize
\begin{tabular}{@{}
  >{\raggedright\arraybackslash}p{(\linewidth - 20\tabcolsep) * \real{0.1000}}
  >{\raggedright\arraybackslash}p{(\linewidth - 20\tabcolsep) * \real{0.0600}}
  >{\raggedright\arraybackslash}p{(\linewidth - 20\tabcolsep) * \real{0.0600}}
  >{\raggedleft\arraybackslash}p{(\linewidth - 20\tabcolsep) * \real{0.0975}}
  >{\raggedleft\arraybackslash}p{(\linewidth - 20\tabcolsep) * \real{0.0975}}
  >{\raggedleft\arraybackslash}p{(\linewidth - 20\tabcolsep) * \real{0.0975}}
  >{\raggedleft\arraybackslash}p{(\linewidth - 20\tabcolsep) * \real{0.0975}}
  >{\raggedleft\arraybackslash}p{(\linewidth - 20\tabcolsep) * \real{0.0975}}
  >{\raggedleft\arraybackslash}p{(\linewidth - 20\tabcolsep) * \real{0.0975}}
  >{\raggedleft\arraybackslash}p{(\linewidth - 20\tabcolsep) * \real{0.0975}}
  >{\raggedleft\arraybackslash}p{(\linewidth - 20\tabcolsep) * \real{0.0975}}@{}}
\toprule\noalign{}
\begin{minipage}[b]{\linewidth}\raggedright
Matchup
\end{minipage} & \begin{minipage}[b]{\linewidth}\raggedright
Hotel 0
\end{minipage} & \begin{minipage}[b]{\linewidth}\raggedright
Hotel 1
\end{minipage} & \begin{minipage}[b]{\linewidth}\raggedleft
RevPAR 0
\end{minipage} & \begin{minipage}[b]{\linewidth}\raggedleft
RevPAR 1
\end{minipage} & \begin{minipage}[b]{\linewidth}\raggedleft
Gap
\end{minipage} & \begin{minipage}[b]{\linewidth}\raggedleft
Occ 0
\end{minipage} & \begin{minipage}[b]{\linewidth}\raggedleft
Occ 1
\end{minipage} & \begin{minipage}[b]{\linewidth}\raggedleft
ADR 0
\end{minipage} & \begin{minipage}[b]{\linewidth}\raggedleft
ADR 1
\end{minipage} & \begin{minipage}[b]{\linewidth}\raggedleft
L1
\end{minipage} \\
\midrule\noalign{}
NC vs NC & NC & NC & 107.403 & 107.389 & +0.014 & 0.7726 & 0.7725 &
139.02 & 139.03 & 0.0054 \\
CA vs CA & CA & CA & 107.488 & 107.477 & +0.010 & 0.7741 & 0.7740 &
138.88 & 138.88 & 0.0018 \\
NC vs CA & NC & CA & 107.553 & 107.963 & -0.410 & 0.7717 & 0.7743 &
139.39 & 139.45 & 0.0310 \\
CA vs NC & CA & NC & 107.820 & 107.388 & +0.432 & 0.7733 & 0.7706 &
139.43 & 139.36 & 0.0294 \\
\end{tabular}
\end{table*}
}

NC-vs-NC and CA-vs-CA are symmetric. In mixed deployment, the CA policy
has a small directional advantage: RevPAR 107.891 versus 107.471,
occupancy 0.7738 versus 0.7711, and ADR 139.44 versus 139.37.

The supported claim is modest: the learned policies do not immediately
collapse back to the low-price shortcut when Fixed RM is removed. We do
not treat the roughly 0.4 RevPAR mixed-matchup gap as a large business
advantage; it is directional evidence that additional lagged
competitor-price context can matter. This is useful deployment evidence,
but not a proof of online co-learning safety.

\section{Robustness and
Boundaries}\label{robustness-and-boundaries}

Aggregate trace alignment can hide local failure. We therefore slice
behavior by early/late horizon and high/low Hotel A inventory.

{\def\LTcaptype{none} 
\begin{table*}[t]
\centering
\scriptsize
\begin{tabular}{@{}
  >{\raggedright\arraybackslash}p{(\linewidth - 10\tabcolsep) * \real{0.1304}}
  >{\raggedleft\arraybackslash}p{(\linewidth - 10\tabcolsep) * \real{0.1739}}
  >{\raggedleft\arraybackslash}p{(\linewidth - 10\tabcolsep) * \real{0.1739}}
  >{\raggedleft\arraybackslash}p{(\linewidth - 10\tabcolsep) * \real{0.1739}}
  >{\raggedleft\arraybackslash}p{(\linewidth - 10\tabcolsep) * \real{0.1739}}
  >{\raggedleft\arraybackslash}p{(\linewidth - 10\tabcolsep) * \real{0.1739}}@{}}
\toprule\noalign{}
\begin{minipage}[b]{\linewidth}\raggedright
Policy
\end{minipage} & \begin{minipage}[b]{\linewidth}\raggedleft
Overall L1
\end{minipage} & \begin{minipage}[b]{\linewidth}\raggedleft
Early high inv.
\end{minipage} & \begin{minipage}[b]{\linewidth}\raggedleft
Early low inv.
\end{minipage} & \begin{minipage}[b]{\linewidth}\raggedleft
Late high inv.
\end{minipage} & \begin{minipage}[b]{\linewidth}\raggedleft
Late low inv.
\end{minipage} \\
\midrule\noalign{}
CA teacher vs Hotel B & 0.0194 & 0.0274 & 0.0236 & 0.0300 & 0.0167 \\
NC student vs Hotel B & 0.0236 & 0.0240 & 0.0640 & 0.0499 & 0.0235 \\
\end{tabular}
\end{table*}
}

The teacher is close across slices. The student has a visible weakness
in early-low-inventory and late-high-inventory slices. A targeted
rare-state repair improved one local slice but worsened global
alignment, so it is not selected. This is a diagnostic value of the
framework: it marks where transfer is weaker, instead of hiding the
issue behind aggregate RevPAR.

We also retrain the full stack across five market variants. This tests
whether the method only works because one default checkpoint was lucky.

{\def\LTcaptype{none} 
\begin{table*}[t]
\centering
\scriptsize
\begin{tabular}{@{}
  >{\raggedright\arraybackslash}p{(\linewidth - 8\tabcolsep) * \real{0.1667}}
  >{\raggedright\arraybackslash}p{(\linewidth - 8\tabcolsep) * \real{0.1667}}
  >{\raggedleft\arraybackslash}p{(\linewidth - 8\tabcolsep) * \real{0.2222}}
  >{\raggedleft\arraybackslash}p{(\linewidth - 8\tabcolsep) * \real{0.2222}}
  >{\raggedleft\arraybackslash}p{(\linewidth - 8\tabcolsep) * \real{0.2222}}@{}}
\toprule\noalign{}
\begin{minipage}[b]{\linewidth}\raggedright
Variant
\end{minipage} & \begin{minipage}[b]{\linewidth}\raggedright
Market change
\end{minipage} & \begin{minipage}[b]{\linewidth}\raggedleft
Teacher L1 vs B
\end{minipage} & \begin{minipage}[b]{\linewidth}\raggedleft
Student L1 vs B
\end{minipage} & \begin{minipage}[b]{\linewidth}\raggedleft
Student RevPAR gap vs teacher
\end{minipage} \\
\midrule\noalign{}
default & original calibration & 0.0211 & 0.0128 & -0.372 \\
low\_demand & lower arrivals & 0.0139 & 0.0143 & -0.338 \\
high\_demand & higher arrivals & 0.0215 & 0.0354 & -1.416 \\
high\_rate\_rm & tighter price floor & 0.0251 & 0.0344 & -0.887 \\
high\_occ\_rm & looser price floor & 0.0349 & 0.0217 & -0.867 \\
\end{tabular}
\end{table*}
}

The method remains strong in the default and low-demand settings. High
demand and changed RM rules expose larger student-transfer gaps. This
should be stated as a boundary, not hidden as noise.

\section{Hidden-Budget Bidding}\label{hidden-budget-bidding}

The hotel simulator is the main testbed. To make the paradigm less
hotel-only, we add a second economic-agent POMDP: hidden-budget bidding.
A bidding policy chooses bid levels for sequential ad opportunities. The
full-state expert observes remaining campaign budget and paces spend.
Partial policies see public quality, competition, time, and lagged bid,
but not the hidden budget.

The formal mapping follows the same discipline-stability paradigm. Let
\(b_t\), \(u_t\), \(c_t\), and \(a_{t-1}\) denote budget, opportunity
quality, competition, and the previous bid. The full state is
\(s_t=(t,b_t,u_t,c_t,a_{t-1})\), while the partial observation is
\(o_t=(t,u_t,c_t,a_{t-1})\). The action is a bid bucket, the outcome is
conversion value per step, and the discipline trace is bid distribution
plus pacing behavior.

We evaluate four variants, each with 10 seeds: default, tight budget,
high competition, and volatile quality. The result is not a full
benchmark suite, but it is stronger second-domain evidence than a single
narrow setting.

Across variants, the full-state budget-pacing expert obtains value/step
0.482 with pacing gap 0.043. The outcome-only aggressive bidder is worse
on both outcome and discipline: value/step 0.374, pacing gap 0.212, L1
0.9859, and JS 0.1905 versus the expert. Partial argmax imitation almost
recovers scalar value (0.475) but remains trace-wrong: L1 0.3549 and JS
0.0394. Partial trace-prior sampling gives a slightly lower value/step
(0.467) but best preserves discipline: pacing gap 0.059, L1 0.0130, and
JS approximately 0.

The key comparison is argmax imitation versus trace-prior sampling.
Argmax nearly matches scalar value, but it collapses the hidden-budget
mixture into a different bid distribution. Trace-prior sampling gives up
a small amount of value while preserving the expert trace.

The bid distributions make this visible. The expert uses bid buckets
\texttt{{[}0.45,\ 0.70,\ 0.95,\ 1.20,\ 1.50{]}} with shares
\texttt{{[}11.59,\ 28.59,\ 42.60,\ 16.55,\ 0.68{]}\%}. Argmax shifts
mass toward the modal 0.95 bucket, producing
\texttt{{[}0.86,\ 26.60,\ 60.34,\ 12.20,\ 0.00{]}\%}. Trace-prior
sampling tracks the expert mixture:
\texttt{{[}11.16,\ 28.62,\ 43.21,\ 16.37,\ 0.64{]}\%}.

Seed-level trace uncertainty tells the same story. Outcome-only
aggressive bidding has seed L1 0.9859 with 95\% CI
\texttt{{[}0.9219,\ 1.0498{]}}; argmax imitation has seed L1 0.4349
\texttt{{[}0.4159,\ 0.4539{]}}; trace-prior sampling has seed L1 0.0171
\texttt{{[}0.0159,\ 0.0184{]}}.

This second domain shows the same failure pattern outside hotels: scalar
outcome can remain close while discipline fails; deterministic argmax
improves outcome but collapses hidden-state uncertainty; stochastic
trace-prior sampling best preserves the expert trace.

We also include a smaller hidden-queue routing check in the appendix
material. It shows the same pattern in a two-pipeline routing task, but
we treat hidden-budget bidding as the main second-domain evidence
because it is closer to strategic economic decision-making.

\section{Baseline Scope}\label{baseline-scope}

For this paper, we do not run MAPPO or MADDPG.

This is a scope decision rather than a claim that these methods are
irrelevant or already covered. The current paper asks whether
scalar-reward learning, memory, training-time critic access, and
trace-access baselines are enough in a fixed-RM benchmark with frozen
deployment tests. It does not ask whether a fully adaptive learned-agent
market can be solved by online co-learning MARL.

The reason is:

\begin{enumerate}
\def\labelenumi{\arabic{enumi}.}
\tightlist
\item
  The primary training environment is a fixed-opponent benchmark, not an
  online co-learning MARL game. Hotel B is Fixed RM.
\item
  The most direct information-symmetry objection has been addressed:
  BC-only, BC warm-start, and PPO + BC auxiliary baselines receive
  Fixed-RM trace labels.
\item
  CTDE PPO tests a narrower question: whether hidden training-time
  information in the critic is enough when the deployed actor still
  optimizes scalar reward. It is not a substitute for MAPPO.
\item
  MAPPO is most relevant if the paper claims cooperative or co-learning
  MARL generality. MADDPG is most relevant for mixed
  cooperative-competitive actor-critic baselines, often with continuous
  control assumptions. Those are broader than the supported claim here.
\item
  Offline RL baselines such as CQL or BCQ are also not direct
  substitutes. They penalize out-of-distribution actions, but they do
  not solve the central issue if the scalar reward itself is the
  incomplete objective.
\end{enumerate}

The paper should therefore state the boundary directly: we do not claim
that all modern MARL algorithms fail. MAPPO/MADDPG-style online
co-learning baselines are future work if the benchmark is extended from
fixed RM plus frozen deployment to fully adaptive learned-agent markets.

This boundary is deliberate. It preserves the contribution as an
evaluation-and-benchmark result rather than a broad negative theorem
about modern MARL.

\section{Related Work}\label{related-work}

Reward hacking and proxy misspecification. Goodhart's Law warns that a
measure can lose validity when optimized as a target (Goodhart, 1975).
Recent ML work formalizes related failure modes as reward hacking and
reward misspecification (Skalse et al., 2022; Pan et al., 2022). Our
setting is a trace-level version of this problem: RevPAR can look
acceptable while behavior fails market discipline.

Revenue management and dynamic pricing. Classical RM studies pricing and
inventory control under stochastic demand (Gallego and van Ryzin, 1994;
Talluri and van Ryzin, 2004). Our contribution is not a better RM
algorithm. We use RM as a controlled economic-agent environment where
the intended behavior is interpretable and the hidden-state trigger is
clear.

POMDPs and hidden-state aliasing. The learner acts under partial
observability (Kaelbling et al., 1998): Hotel B's inventory is hidden.
Our oracle-\(q_B\) diagnostic makes this mechanism empirical
rather than merely conceptual.

Imitation learning and DAgger. The student-transfer stage uses a
DAgger-style correction idea (Ross et al., 2011): the teacher labels the
states induced by the student, not only the original teacher
trajectories. Our use is pragmatic and benchmark-facing. We also include
information-symmetric behavior-cloning baselines because a method that
receives a market trace should not be compared only to reward-only RL.
The result is deliberately nuanced: BC-only shows that the trace is
learnable, while BC warm-start and sampled-action BC auxiliaries show
that trace access alone is not the same as a full-distribution
discipline prior inside RL.

KL-regularized and prior-regularized RL. Trace-Prior RL is related to
KL-regularized policy optimization and policy regularization, including
MPO, Distral, BRAC, AWAC, and KL-control-style dialogue RL (Teh et
al., 2017; Abdolmaleki et al., 2018; Jaques et al., 2019; Wu et
al., 2020; Nair et al., 2020). We do not claim a new optimizer. The key
difference is the role of the prior: it estimates a competitor/benchmark
trace distribution under hidden state, and the evaluation target is
market-discipline alignment rather than safe improvement over the
agent's own offline behavior.

MARL baselines. PPO, recurrent PPO, and CTDE PPO are included because
they directly answer whether stronger reward-only policy optimization,
memory, or centralized critic information solves the benchmark. MAPPO
(Yu et al., 2022) and MADDPG (Lowe et al., 2017) remain future work for
a fully online co-learning version.

Offline RL. Conservative offline RL methods are important when
deployment actions may leave the support of the dataset. They are not
the primary baseline for this paper because the failure is not only
out-of-distribution action selection. The deeper problem is objective
incompleteness: a scalar reward can be optimized while the intended
trace discipline fails.

\section{Limitations}\label{limitations}

The simulator is synthetic. We do not claim real hotel validity without
calibration to real booking and pricing data.

Hotel B is Fixed RM. This is deliberate for causal clarity, but it is
not a living strategic competitor. If Hotel B were also learning,
discipline stability could degrade because the benchmark trace would
become non-stationary. A natural extension is to estimate a moving or
ensemble trace prior rather than a single fixed prior.

The trace prior is only as good as the benchmark trace. If the available
market trace is noisy, biased, collusive, or commercially undesirable,
Trace-Prior RL will preserve that flawed discipline. The method
therefore requires a benchmark selection step: decide which trace is
worth preserving before using it as a prior.

BC-only is a strong baseline in the default symmetric market. This
limits the algorithmic claim: the main contribution is the
discipline-stability benchmark and repair ladder, not proof that
KL-regularized RL always improves on direct trace imitation. The
capacity-asymmetric \(Q_A=120\) test gives the cleanest evidence that RL
can add objective adaptation beyond BC-only, but the gain is still modest
and appears through occupancy rather than ADR. That is, Trace-Prior RL
fills more rooms at similar rates rather than charging more per room,
which is the natural way to use additional capacity. The smaller
cost-based and low-capacity asymmetry tests remain boundary evidence
rather than headline results.

Hidden-budget bidding is a compact synthetic second domain, not a mature
benchmark suite or real ad platform. The hidden-queue routing check is
supporting appendix-style evidence.

The student transfer is strong but not perfect. Rare-state slices and
high-demand variants reveal where the current student needs richer data
or targeted correction.

Frozen learned-agent deployment is not the same as online co-learning
safety. MAPPO/MADDPG-style baselines are appropriate if the benchmark is
extended to adaptive learned-agent markets.

\section{Conclusion}\label{conclusion}

The central lesson is simple: outcome is not discipline. A pricing agent
can optimize a local revenue metric while failing to learn the market
behavior that makes the policy deployable. In the pricing benchmark,
this failure is triggered by hidden competitor state: Hotel A cannot observe the
inventory variable that Hotel B uses to price.

The paradigm is also simple at the conceptual level. First, define the
discipline the benchmark is meant to preserve. Second, map what the
deployed agent can and cannot observe. Third, use failure traces to
define what must be measured. Fourth, learn the benchmark trace as a
distribution, not a single label. Fifth, test whether the learned
discipline transfers and remains stable when the original benchmark
signal is reduced or removed.

This turns the project from ``can Hotel A get good RevPAR?'' into a
broader evaluation question for strategic economic agents:

\begin{quote}
Does the agent preserve the behavioral structure of the system it
enters?
\end{quote}

That question is useful for revenue management, routing, bidding, and
legal or compliance audits of pricing systems. For industry, the lesson
is that a pricing agent should be evaluated by its traces, not only by a
KPI dashboard. For legal or litigation support, the same diagnostics can
help describe what a pricing system did: whether it matched a benchmark
distribution, undercut in specific states, or changed the market path it
was learning from. These diagnostics do not prove intent or liability.
They provide reproducible behavioral evidence that can separate ordinary
discipline, shortcut optimization, and market-distorting traces.

{\small
\nocite{*}
\bibliographystyle{plainnat}
\bibliography{references}
}

\end{document}